%% file: ArXiv_paper.tex
\documentclass[10pt,twocolumn,letterpaper]{article}

\pdfoutput=1

\usepackage{cvpr}
\usepackage{times}
\usepackage{epsfig}
\usepackage{graphicx}
\usepackage{amsmath}
\usepackage{amssymb}

\usepackage{color}
\usepackage[utf8]{inputenc}
\usepackage{stmaryrd}
\usepackage{algorithm}
\usepackage[noend]{algpseudocode}
\usepackage{float}
\usepackage{subfig}
\usepackage{flushend}

\usepackage{soul}
\usepackage[mathcal]{euscript}
\usepackage{sidecap}
\usepackage{multirow}
\usepackage{booktabs}
\usepackage{authblk}

 \cvprfinalcopy 

\begin{document}

\title{Approximate search with quantized sparse representations}
\author[1,2]{Himalaya Jain \thanks{himalaya.jain@inria.fr}}
\author[2]{Patrick P\'erez \thanks{patrick.perez@technicolor.com}}
\author[1]{R\'emi Gribonval \thanks{remi.gribonval@inria.fr}}
\author[2]{Joaquin Zepeda\thanks{joaquin.zepeda@technicolor.com}}
\author[1]{Herv\'e J\'egou \thanks{rvj@fb.com}}
\affil[1]{Inria Rennes, France}
\affil[2]{Technicolor, France}

\maketitle
\input{macro.tex}

\begin{abstract}
    This paper tackles the task of storing a large collection of vectors, such as visual descriptors, and of searching in it.
    To this end, we propose to approximate database vectors by constrained sparse coding, where possible atom weights are restricted to belong to a finite subset.
    This formulation encompasses, as particular cases, previous state-of-the-art methods such as product or residual quantization.
    As opposed to traditional sparse coding methods, quantized sparse coding includes 
    memory usage as a design constraint, thereby allowing us to index a large collection such as the BIGANN billion-sized benchmark.
    Our experiments, carried out on standard benchmarks, show that our formulation leads to competitive solutions when considering different trade-offs between learning/coding time, index size and search quality.
\end{abstract}

\input{introduction2.tex}
\input{related2.tex}
\input{approach2.tex}

\input{ann_search2.tex}
\input{experiments_arxiv.tex}

{\small
	\bibliographystyle{ieee}
	\bibliography{ArXiv_paper}
}

\end{document}

%% file: macro.tex
\newfloat{algorithm}{t}{lop}

\def \argmin  {\arg\!\min}
\def \argmax  {\arg\!\max}

\newcommand{\mbf}[1]{\ensuremath{\mathbf{#1}}}
\newcommand{\mbs}[1]{\ensuremath{\boldsymbol{#1}}}
\newcommand{\mcl}[1]{\ensuremath{\mathcal{#1}}}
\newcommand{\mrm}[1]{\ensuremath{\mathrm{#1}}}
\newcommand{\mbb}[1]{\ensuremath{\mathbb{#1}}}

\newcommand{\bigo}[1]{\ensuremath{\mathcal{O}(#1)}}
\newcommand{\trsp}[1]{\ensuremath{#1^{\top}}}
\newcommand{\trace}[1]{\ensuremath{\mathrm{trace}(#1)}}
\newcommand{\deter}[1]{\ensuremath{\mathrm{det}(#1)}}
\newcommand{\pinv}[1]{\ensuremath{#1^{\dagger}}}
\newcommand{\diag}[1]{\ensuremath{\mathrm{diag}(#1)}}
\newcommand{\rk}[1]{\ensuremath{\mathrm{rank}(#1)}}
\newcommand{\bmat}[4]{\ensuremath{\begin{bmatrix}#1&#2\\#3&#4\end{bmatrix}}}

\def\ie{\emph{i.e.}}
\def\eg{\emph{e.g.}}
\def\wrt{\emph{w.r.t.~}}
\def\mwrt{\mrm{w.r.t.~}}
\def\msbt{\mrm{sb.t.~}}

\def\cX{\mcl{X}}
\def\cY{\mcl{Y}}
\def\cZ{\mcl{Z}}

\def\bc{\mbf{c}}
\def\tbc{\tilde{\mbf{c}}}
\def\bq{\mbf{q}}
\def\ba{\mbf{a}}
\def\by{\mbf{y}}
\def\bz{\mbf{z}}
\def\tbz{\tilde{\mbf{z}}}
\def\bx{\mbf{x}}
\def\tbx{\tilde{\mbf{x}}}
\def\br{\mbf{r}}
\def\bk{\mbf{k}}
\def\balpha{\mbs{\alpha}}

\def\L2n{$\ell^2$-norm}

\def\K{\llbracket 1,K \rrbracket}
\def\P{\llbracket 1,P \rrbracket}

\def\aPQ{\mbox{$\alpha$-PQ}}
\def\qaPQ{\mbox{Q$\alpha$-PQ}}
\def\aRQ{\mbox{$\alpha$-RVQ}}
\def\qaRQ{\mbox{Q$\alpha$-RVQ}}

%% file: introduction2.tex
\section{Introduction}\label{sec:into}

Retrieving, from a very large database of high-dimensional vectors, the ones that ``resemble" most a query vector is at the heart of most modern information retrieval systems. Online exploration of very large media repositories, for tasks ranging from copy detection to example-based search and recognition, routinely faces this challenging problem. Vectors of interest are abstract representations of the database documents that permit meaningful comparisons in terms of distance and similarity. Their dimension typically ranges from a few hundreds to tens of thousands. In visual search, these vectors are ad-hoc or learned descriptors that represent image fragments or whole images.

Searching efficiently among millions or billions of such high-dimensional vectors requires specific techniques. The classical approach is to re-encode all vectors in a way that allows the design of a compact index and the use of this index to perform fast \textit{approximate search} for each new query. Among the different encoding approaches that have been developed for this purpose, state-of-the-art systems rely on various forms of vector quantization: database vectors are approximated using compact representations that can be stored and searched efficiently, while the query need not be approximated (asymmetric approximate search). In order to get high quality approximation with practical complexities, the encoding is structured, typically expressed as a sum of codewords stemming from suitable codebooks. There are two 
main classes of such structured quantization techniques: those based on vector partitioning and independent quantization of sub-vectors \cite{GHKS13,JDS11,NF13}; those based on sequential residual encoding \cite{AYWHG15,CGW10,jegou2011searching,MHL14,ZDW14,ZQTW15}.

In this work, we show how
these approaches can be taken one step further by drawing inspiration from the sparse coding interpretation of these techniques~\cite{VZ12a}. The key idea is to represent input vectors as linear combinations of atoms, instead of sums of codewords. The introduction of scalar weights allows us to extend both residual-based and partitioned-based quantizations such that approximation quality is further improved with modest overhead. For this extension to be compatible with large scale approximate search, the newly introduced scalar weights must be themselves encoded in a compact way. We propose to do so by quantizing the vector they form. The resulting scheme will thus trade part of the original encoding budget for encoding coefficients. As we shall demonstrate on various datasets, the proposed quantized sparse representation (i) competes with partitioned quantization for equal memory footprint and lower learning/coding complexity and (ii) outperforms residual quantization with equal 
or smaller memory footprint and learning/coding complexity.

In the next section, we discuss in more details the problem of approximate vector search with structured quantizers and recall useful concepts from sparse coding. With these tools in hand, we introduce in Section \ref{sec:approach} the proposed structured encoding by quantized sparse representations. The different bricks --learning, encoding and approximate search-- are presented in Sections \ref{sec:sparse} and \ref{sec:search}, both for the most general form of the framework (residual encoding with non-orthogonal dictionaries) and for its partitioned variant. Experiments are described and discussed in Section \ref{sec:experiments}.

%% file: related2.tex
\section{Related work}\label{sec:related}

Approximate vector search is a long-standing research topic across a wide range of domains, from communication and data mining to computer graphics and signal processing, analysis and compression. Important tools have been developed around hashing techniques \cite{IM98}, which turn the original search problem into the one of comparing compact codes, \ie, binary codes~\cite{C02}, see \cite{wangLKC15} for a recent overview on binary hashing techniques. Among other applications, visual search has been addressed by a number of such binary encoding schemes (\eg, \cite{lv2004image,NPF12,torralba2008small,WKC12,XWLZLY11}).

An important aspect of hashing and related methods is that their efficiency comes at the price of comparing only codes and not vectors in the original input space. In the present work we focus on another type of approaches that are currently state-of-art in large scale visual search. Sometimes referred to as \textit{vector compression} techniques, they provide for each database vector $\bx$ an approximation $Q(\bx)\approx \bx$ such that (i) the Euclidean distance (or other related similarity measure such as inner product or cosine) to any query vector $\by$ is well estimated using $Q(\bx)$ instead of $\bx$ and (ii) these approximate (di)similarity measures can be efficiently computed using the code that defines $Q(\bx)$.

A simple way to do that is to rely on vector quantization \cite{gersho2012vector}, which maps $\bx$ to the closest vector in a codebook learned through $k$-means clustering. In high dimensions though, the complexity of this approach grows to maintain fine grain quantization. A simple and powerful way to circumvent this problem is to partition vectors into smaller dimensional sub-vectors that are then vector quantized. At the heart of product quantization (PQ) \cite{JDS11}, this idea has proved very effective for approximate search within large collections of visual descriptors. Different extensions, such as ``optimized product quantization'' (OPQ) \cite{GHKS13} and ``Cartesian $k$-means" (CKM) \cite{NF13} optimize the chosen partition, possibly after rotation, such that the distortion $\|\bx-Q(\bx)\|$ is further reduced on average. Additionally, part of the encoding budget can be used to encode this distortion and improve the search among product-quantized vectors \cite{HLY14}.

It turns out that this type of partitioned quantization is a special case of \textit{structured} or layered quantization:
\begin{equation}
Q(\bx) = \sum_{m=1}^M Q_m(\bx),
\label{eq:ac}
\end{equation}
where each quantizer $Q_m$ uses a specific codebook. In PQ and its variants, these codebooks are orthogonal, which makes learning, encoding and search especially efficient. Sacrificing part of this efficiency by relaxing the orthogonality constraint can nonetheless provide better approximations. A number of recent works explore this path.

``Additive quantization" (AQ) \cite{BL14} is probably the most general of those, hence the most complex to learn and use. It indeed addresses the combinatorial problem of jointly finding the best set of $M$ codewords in \eqref{eq:ac}. While excellent approximation and search performance is obtained, its high computational cost makes it less scalable \cite{babenko2015tree}. In particular, it is not adapted to the very large scale experiments we report in present work. 

In ``composite quantization" (CQ) \cite{ZDW14}, the overhead caused at search time by the non-orthogonality of codebooks is alleviated by learning codebooks that ensure $\|Q(\bx)\|= \mathrm{cst}$. This approach can be sped up by enforcing in addition the sparsity of codewords \cite{ZQTW15}. As AQ --though to a lesser extent-- CQ and its sparse variant have high learning and encoding complexities.

A less complex way to handle sums of codewords from non-orthogonal codebooks is offered by the greedy approach of ``residual vector quantization" (RVQ) \cite{BRN96,JH82}. The encoding proceeds sequentially such that the $m$-th quantizer encodes the \textit{residual} $\bx-\sum_{n=1}^{m-1}Q_{n}(\bx)$. Accordingly, codebooks are also learned sequentially, each one based on the previous layer's residuals from the training set. This classic vector quantization approach was recently used for approximate search \cite{AYWHG15,CGW10,MHL14}.  
``Enhanced residual vector quantization" (ERVQ) \cite{AYWHG15} improves the performance by jointly refining the codebooks in a final training step, while keeping purely sequential the encoding process.

Important to the present work, \textit{sparse coding} is another powerful way to approximate and compress vectors \cite{elad2010sparse}. In this framework, a vector is also approximated as in \eqref{eq:ac}, but with each $Q_m(\bx)$ being of the form $\alpha_m \bc_{k_m}$, where $\alpha_m$ is a scalar weight and $\bc_{k_m}$ is a unit norm \textit{atom} from a learned \textit{dictionary}. The number of selected atoms can be pre-defined or not, and these atoms can stem from one or multiple dictionaries. A wealth of techniques exist to learn dictionaries and encode vectors \cite{elad2010sparse,mairal2010online,wright2010sparse}, including ones that use the Cartesian product of sub-vector dictionaries \cite{GHS14} similarly to PQ  or residual encodings \cite{zepeda2011jstsp,zepeda2010mmsp} similarly to RQ to reduce encoding complexity. Sparse coding thus offers representations that are related to structured quantization, and somewhat richer. Note however that these representations are not discrete in general, which makes them \textit{a priori} ill-suited to indexing very 
large vector collections. Scalar quantization of the weights has nonetheless been proposed in the context of audio and image compression \cite{zepeda2011jstsp,frossard2004posteriori,yaghoobi2007quantized}.

Our proposal is to import some of the ideas of sparse coding into the realm of approximate search. In particular, we propose to use sparse representations over possibly non-orthogonal dictionaries and with vector-quantized coefficients, which offer interesting extensions of both partitioned and residual quantizations.

%% file: approach2.tex
\section{Quantized sparse representations}\label{sec:approach}
 
\textbf{A sparse coding view of structured quantization} \quad  
Given $M$ codebooks, structured quantization represents each database vector $\bx$ as a sum~\eqref{eq:ac} of $M$ codewords, one from each codebook. Using this decomposition, search can be expedited by working at the atom level (see Section \ref{sec:search}).
Taking a sparse coding viewpoint, we propose a more general approach whereby $M$ dictionaries\footnote{Throughout we use the terminology \textit{codebook} for a collection of vectors, the \textit{codewords}, that can be added, and \textit{dictionary} for a collection of normalized vectors, the \textit{atoms}, which can be linearly combined.}, 
$C^m = [\bc^m_1\cdots\bc^m_K]_{D\times K}$, $m=1\cdots M$, each with $K$ normalized atoms, are learned and a database vector $\bx\in\mathbb{R}^D$ is represented as a linear combination:
\begin{equation}
	Q(\bx) = \sum_{m=1}^M \alpha_m(\bx) \bc^m_{k_m(\bx)},
	\label{eq:coding}
\end{equation}
where $\alpha_m(\bx)\in\mathbb{R}$ and $k_{m}(\bx)\in\K$.
Next, we shall drop the explicit dependence in $\bx$ for notational convenience.  
As we shall see in Section \ref{sec:experiments} (Fig. \ref{fig:dst}), the additional degrees of freedom provided by the weights in (\ref{eq:coding}) allow more accurate vector approximation.
However, with no constraints on the weights, this representation is not discrete, spanning 
a union of $M$-dimensional sub-spaces in $\mathbb{R}^{D}$. To produce compact codes, it must be restricted. Before addressing this point, we show first how it is obtained and how it relates to existing coding schemes.

If dictionaries are given, trying to compute $Q(\bx)$ as the best \L2n approximation of $\bx$ is a special case of sparse coding, with the constraint of using exactly one atom from each dictionary. Unless dictionaries are mutually orthogonal, it is a combinatorial problem that can only be solved approximately. Greedy techniques such as projection pursuit \cite{friedman1981projection} and matching pursuit \cite{mallat1993matching} provide particularly simple ways to compute sparse representations. We propose the following pursuit for our problem: for $m = 1\cdots M$,
\begin{equation}
   k_{m} =  \argmax_{k\in\K} \trsp{\br_m}\bc_k^{m},~~
   \alpha_{m} =  \trsp{\br_m}\bc_{k_{m}}^{m},
   \label{eq:pursuit}
\end{equation}
with $\br_1=\bx$ and $\br_{m+1} = \br_{m} - \alpha_{m}\bc^m_{k_m}$. Encoding proceeds recursively, selecting in the current dictionary the atom with maximum inner-product with the current residual.\footnote{\emph{Not} maximum absolute inner-product as in matching pursuit. This permits to get a tighter distribution of weights, which will make easier their subsequent quantization.}
Once atoms have all been sequentially selected, \ie, the support of the $M$-sparse representation is fixed, the approximation (\ref{eq:coding}) is refined by jointly recomputing the weights as
\begin{equation}
	\hat{\balpha} = \argmin_{\balpha\in\mathbb{R}^M} \|\bx - C(\bk)\balpha\|_2^2 = \pinv{C(\bk)}\bx,
\label{eq:alpha}
\end{equation}
with $\bk = [k_m]_{m=1}^M \in \K^{M}$ the vector formed by the selected atom indices, $C(\bk) = [\bc^1_{k_1}\cdots \bc^M_{k_M}]_{D \times M}$ the corresponding atom matrix and $(\cdot)^{\dagger}$ the Moore-Penrose pseudo-inverse.
Vector $\hat{\balpha}$ contains the $M$ weights, out of $KM$, associated to the selected support. Note that the proposed method is related to \cite{zepeda2011jstsp,zepeda2010mmsp}.\\

\noindent\textbf{Learning dictionaries} \quad
In structured vector quantization, the $M$ codebooks are learned on a limited training set, usually through $k$-means. In a similar fashion, $k$-SVD on a training set of vectors is a classic way to learn dictionaries for sparse coding \cite{elad2010sparse}. In both cases, encoding of training vectors and optimal update of atoms/codewords alternate until a criterion is met, starting from a sensible initialization (\eg, based on a random selection of training vectors). Staying closer to the spirit of vector quantization, we also rely on $k$-means in its spherical variant which fits well our needs: spherical $k$-means iteratively clusters vector \textit{directions}, thus delivering meaningful unit atoms.

Given a set $\cZ = \{\bz_1\cdots \bz_R\}$ of $R$ training vectors, the learning of one dictionary of $K$ atoms proceeds iteratively according to:
\begin{align}
	\mrm{Assignment:~} & k_r = \arg\max_{k\in\K} \trsp{\bz_r}{\bc_k},~ \forall r\in\llbracket 1,R \rrbracket, \\
	\mrm{Update:~}&  \bc_k \propto\!\!\!\! \sum_{r:k_r=k} \!\!\! \bz_r,~\|\bc_k\|=1,~\forall k\in\K.
\end{align}
This procedure is used to learn the $M$ dictionaries. The first dictionary is learned on the training vector themselves, the following ones on corresponding residual vectors.  However, in the particular case where dictionaries are chosen within prescribed mutually orthogonal sub-spaces, they can be learned independently after projection in each-subspace, as discussed in Section \ref{sec:sparse}.\\

\noindent\textbf{Quantizing coefficients}\quad
To use the proposed representation for large-scale search,
we need to limit the possible values of coefficients while maintaining good approximation quality.
Sparse representations with discrete weights have been proposed in image and audio compression \cite{frossard2004posteriori,yaghoobi2007quantized},
however with scalar coefficients that are quantized independently and not in the prospect of approximate search.
We propose a novel approach that serves our aim better, namely employing vector quantization of coefficient vectors $\hat{\balpha}$.
These vectors are of modest size, \ie, $M$ is between 4 and 16 in our experiments. Classical $k$-means clustering is thus well adapted to produce a codebook $A=[\ba_1\cdots\ba_P]_{M\times P}$ for their quantization. This is done after the main dictionaries have been learned.\footnote{Alternate refinement of the vector dictionaries $C^m$s and of the coefficient codebook $A$ led to no improvement. 
A possible reason is that dictionaries update does not take into account that the coefficients are vector quantized, and we do not see a principled way to do so. 
}

Denoting $p(\balpha)=\argmin_{p\in\P} \|\balpha-\ba_p\|$ the index of the vector-quantization of $\balpha$ with this codebook, the final approximation of vector $\bx$ reads:
\begin{equation}
	Q(\bx) = C(\bk) \ba_{p(\hat{\balpha})},
	\label{eq:qarq}
\end{equation}
with $\bk$ function of $\bx$ (Eq. \ref{eq:pursuit}) and  $\hat{\balpha}= \pinv{C(\bk)}\bx$ (Eq. \ref{eq:alpha}) function of $\bk$ and $\bx$.\\

\noindent\textbf{Code size}\quad
A key feature of structured quantization is that it provides the approximation accuracy of extremely large codebooks while limiting learning, coding and search complexities: The $M$ codebooks of size $K$ are as expensive to learn and use as a single codebook of size $MK$ but give effectively access to $K^M$ codewords. In the typical setting where $M=8$ and $K=256$, the effective number of possible encodings is $2^{64}$, that is more than $10^{19}$. This 64-bit encoding capability is obtained by learning and using only $8$-bit quantizers. Similarly, quantized sparse coding offers up to $K^M\times P$ encoding vectors, which amounts to $M\log_2K+\log_2 P$ bits. Structured quantization with $M+1$ codebooks, all of size 
$K$ except one of size $P$ has the same code-size, but 
leads to a different discretization of the ambient vector space $\mathbb{R}^{D}$. 
The aim of the experiments will be to understand how trading part of the vector encoding budget for encoding jointly the scalar weights can benefit approximate search.

%% file: ann_search2.tex
\section{Sparse coding extension of PQ and RVQ}\label{sec:sparse}

In the absence of specific constraints on the $M$ dictionaries, the proposed quantized sparse coding can be seen as a generalization of residual vector quantization (RVQ), with linear combinations rather than only sums of centroids. Hierarchical code structure and search methods (see Section~\ref{sec:search} below) are analog. To highlight this relationship, 
we will denote ``\qaRQ " the proposed encoder. 

In case dictionaries are constrained to stem from predefined orthogonal sub-spaces $V_{m}$s such that $\mathbb{R}^D=\bigoplus_{m=1}^M V_m$, the proposed approach simplifies notably. Encoding vectors and learning dictionaries can be done independently in each subspace, instead of in sequence. 
In particular, when each subspace is spanned by $D/M$ (assuming $M$ divides $D$) successive canonical vectors, \eg, $V_1 = \mrm{span}(\mbf{e}_1\cdots\mbf{e}_{D/M})$, our proposed approach is similar to product quantization (PQ), which it extends through the use of quantized coefficients. We will denote ``\qaPQ" our approach in this specific set-up: all vectors are partitioned into $M$ sub-vectors of dimension $D/M$ and each sub-vector is approximated independently, with one codeword in PQ, with the multiple of one atom in \qaPQ. \\

\begin{algorithm}
	\small
	\caption{Learning \qaRQ}
	\label{alg:qaRQ_learn}
\begin{algorithmic}[1]
	\State Input: $\bz_{1:R}$
	\State Ouput: $C^{1:M},A$
	\State $\br_{1:R} \leftarrow \bz_{1:R}$        
	\For {$m=1\cdots M$}
		\State $C^m \leftarrow $ {\sc Spher\_k-Means}($\br_{1:R}$)
        \For{$r=1\cdots R$}
			\State $k_{m,r} \leftarrow  \argmax_{k\in\K} \trsp{\br_r}\bc_k^{m}$         	
        	\State $\br_r \leftarrow \br_r - (\trsp{\br_r}\bc_{k_{m,r}}^{m}) \bc^m_{k_{m,r}}$        			\EndFor        
    \EndFor
    \For{$r=1\cdots R$}
    	\State $\balpha_r \leftarrow \pinv{[\bc^1_{k_{1,r}}\cdots \bc^M_{k_{M,r}}]}\bz_r$
    \EndFor
	\State $A \leftarrow$ {\sc k-Means}($\balpha_{1:R}$)
\end{algorithmic}
\end{algorithm}

\begin{algorithm}
	\small
	\caption{Vector encoding with \qaRQ}
	\label{alg:qaRQ}
\begin{algorithmic}[1]
	\State Input: $\bx,[\bc^{1:M}_{1:K}],[\ba_{1:P}]$
	\State Output: $\bk=[k_{1:M}],p$
	\State $\br \leftarrow \bx$    
    \For {$m=1\cdots M$}
        \State $k_{m} \leftarrow  \argmax_{k\in\K} \trsp{\br}\bc_k^{m}$
        \State $\br \leftarrow \br - (\trsp{\br}\bc_{k_{m}}^{m}) \bc^m_{k_m}$
	\EndFor
	\State $\balpha \leftarrow \pinv{[\bc^1_{k_1}\cdots \bc^M_{k_M}]}\bx$
	\State $p\leftarrow \argmin_{p\in\P} \|\balpha - \ba_p\|$
\end{algorithmic}
\end{algorithm}

\begin{algorithm}
	\small
	\caption{Learning \qaPQ}
	\label{alg:qaPQ_learn}
\begin{algorithmic}[1]
	\State Input: $\bz_{1:R}$
	\State Output: $\tilde{C}^{1:M},A$
	\For{$r=1\cdots R$}
		\State $[\trsp\tbz_{1,r} \cdots \trsp\tbz_{M,r}] \leftarrow \trsp{\bz_{r}}$        
	\EndFor
	\For {$m=1\cdots M$}
		\State $\tilde{C}^m \leftarrow $ {\sc Spher\_k-Means}($\tbz_{m,1:R}$)
        \For{$r=1\cdots R$}
			\State $k\leftarrow  \argmax_{k\in\K} \trsp{\tbz_{m,r}}\tbc_k^{m}$         	
        	\State $\alpha_{m,r} \leftarrow  \trsp{\tbz_{m,r}}\tbc_{k}^{m}$
		\EndFor        
    \EndFor
	\State $A \leftarrow$ {\sc k-Means}($\balpha_{1:R}$)
\end{algorithmic}
\end{algorithm}

\begin{algorithm}
	\small
	\caption{Vector encoding with \qaPQ}
	\label{alg:qaPQ}
\begin{algorithmic}[1]
	\State Input: $\bx,[\tbc^{1:M}_{1:K}],[\ba_{1:P}]$
	\State Output: $\bk=[k_{1:M}],p$	
	\State $[\trsp{\tbx_1}\cdots \trsp{\tbx_M}] \leftarrow \trsp{\bx}$    
    \For {$m=1\cdots M$}
        \State $k_{m} \leftarrow  \argmax_{k\in\K} \trsp{\tbx_m}\tbc_k^{m}$
        \State $\alpha_{m} \leftarrow  \trsp{\tbx_m}\tbc_{k_{m}}^{m}$
	\EndFor
	\State $p\leftarrow \argmin_{p\in\P} \|\balpha - \ba_p\|$
\end{algorithmic}
\end{algorithm}

Learning the dictionaries $C^m$s and the codebook $A$ for \qaRQ~is summarized in Alg. 1, and the encoding of a vector with them is in  Alg. 2. 
Learning and encoding in the product case (\qaPQ) are respectively summarized in Algs. 3 and 4, where all training and test vectors are partitioned in $M$ sub-vectors of dimension $D/M$, denoted with tilde. 

\section{Approximate search}\label{sec:search}

Three related types of nearest neighbor (NN) search are used in practice, depending on how the (dis)similarity between vectors is measured in $\mathbb{R}^D$: minimum Euclidean distance, maximum cosine-similarity or maximum inner-product. The three are equivalent when all vectors are $\ell^2$-normalized. In visual search, classical descriptors (either at local level or image level) can be normalized in a variety of ways, \eg, $\ell^2$, $\ell^1$ or blockwise $\ell^2$, exactly or approximately.

With cosine-similarity (CS) for instance, the vector closest the query $\by$ in the database $\cX$ is $\arg\max_{\bx\in\cX}\frac{\trsp{\by}\bx}{\|\bx\|}$,
where the norm of the query is ignored for it has no influence on the answer. Considering approximations of database vectors with existing or proposed methods, approximate NN (aNN) search can be conducted without approximating the query (asymmetric aNN \cite{JDS11}):
\begin{equation}
	\mrm{CS-aNN:~~} \arg\max_{\bx\in\cX}\frac{\trsp{\by}Q(\bx)}{\|Q(\bx)\|}.
	\label{eq:cs_ann}
\end{equation}

As with structured encoding schemes, the form of the approximation in \eqref{eq:qarq} permits to expedite the search. Indeed, for $\bx$ encoded by $(\bk,p)\in \K^M \times \llbracket 1,P \rrbracket$, 
the approximate cosine-similarity reads
\begin{equation}
	\frac{\trsp{\by}C(\bk)\ba_p}{\|C(\bk)\ba_p\|}, 
	\label{eq:qaRQ_search}
\end{equation}
where the $M$ inner products in $\trsp{\by}C(\bk)$ are among the $MK$ ones in $\trsp{\by}C$, which can be computed once and stored for a given query. For each database vector $\bx$, computing the numerator then requires $M$ look-ups, $M$ multiplications and $M-1$ sums. We discuss the denominator below.

In the \qaPQ~setup, as the $M$ unit atoms involved in $C(\bk)$ are mutually orthogonal, the denominator is equal to $\|\ba_p\|$, that is one among $P$ values that are independent of the queries and simply pre-computed once for all. In $\qaRQ$ however, as in other 
quantizers with non-orthogonal codebooks, the computation of 
\begin{equation}
	\|C(\bk)\ba_p\| = \Big(\sum_{m,n=1}^M a_{mp}a_{np}\bc^{m\top}_{k_m}\bc^n_{k_n}\Big)^{1/2}
\end{equation} 
constitutes an overhead. Two methods are suggested in \cite{BL14} to handle this problem. The first one consists in precomputing and storing for look-up all inter-dictionary inner products of atoms,  
\ie\, $\trsp{C}C$. 
For a given query, the denominator can then be computed with $\bigo{M^2}$ operations. The second method is to compute the norms for all approximated database vectors and to encode them with a non-uniform scalar quantizer (typically with 256 values) learned on the training set. This adds an extra byte to the database vector encoding but avoids the search time overhead incurred by the first method. This computational saving is worth the memory expense for very large scale systems (See experiments on 1 billion vectors in the next section).  	

Using the Euclidean distance instead of the cosine similarity, \ie, solving 
$\arg\min_{\bx\in\cX} \left\{\|Q(\bx)\|^2 - 2~\trsp{\by}Q(\bx)\right\}$ leads to very similar derivations.
The performance of the proposed framework is equivalent for these two popular metrics. \medskip

%% file: experiments_arxiv.tex
\section{Experiments}\label{sec:experiments}
We compare on various datasets the proposed methods, $\qaRQ$ and $\qaPQ$, to the structured quantization techniques they extend, RVQ and PQ respectively.
We use three main datasets: SIFT1M \cite{JDS09}, GIST1M \cite{JDS11} and VLAD500K \cite{AZ13}.\footnote{VLAD vectors, as kindly provided by Relja Arandjelovi\'c, are PCA-compressed to 128 dimensions and unit $\ell^2$-normalized; SIFT vectors are 128-dimensional and have almost constant $\ell^2$-norm of 512, yielding almost identical nearest-neighbors for cosine similarity and $\ell^2$ distance.}
For PQ and $\qaPQ$ on GIST and VLAD vectors, PCA rotation and random coordinate permutation are applied, as they have been shown to improve performance in previous works. Each dataset includes a main set to be searched ($\cX$ of size $N$), a training set ($\cZ$ of size $R$) and $S$ query vectors. These sizes and input dimension $D$ are as follows: \medskip

\centerline{\small
\begin{tabular}{c r r r r}
\toprule
Dataset 	&  $D$ 	& $R$ 	& $N$ 	& $S$ 	\\ \midrule
SIFT1M 		& 	\quad 128	& \quad 100K 	& \quad 1M 	& \quad 10K	\\ 
GIST1M		& 	960	& 500K 	& 1M	& 1K 	\\ 
VLAD500K	& 	128	& 400K	& 0.5M	& 1K    \\ \bottomrule
\end{tabular}}
~

As classically done, we report performance in terms of recall@{\tt R}, \ie, the proportion of query vectors for which the true nearest neighbor is present among the {\tt R} nearest neighbors returned by the approximate search. \medskip

\begin{figure*}	
	\centering
	\scriptsize	
	\begin{tabular}{ccc}	
		\includegraphics[trim = 0mm 3mm 0mm 1mm, clip, width=.33\textwidth]{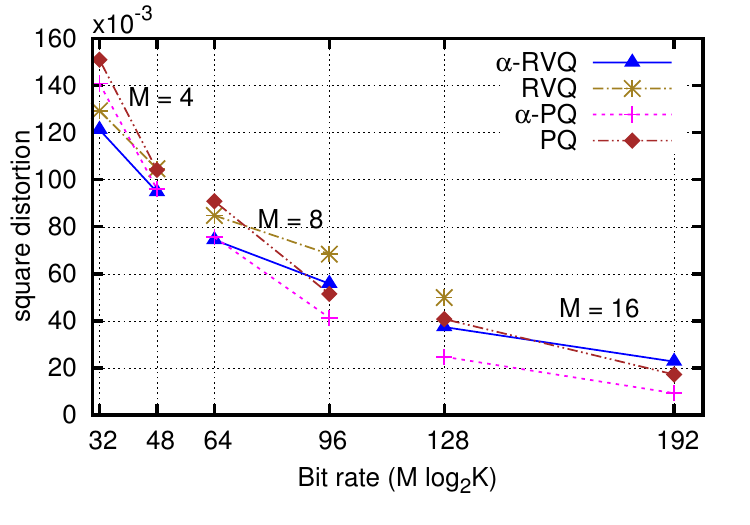}&
		\includegraphics[trim = 0mm 3mm 0mm 1mm, clip, width=.33\textwidth]{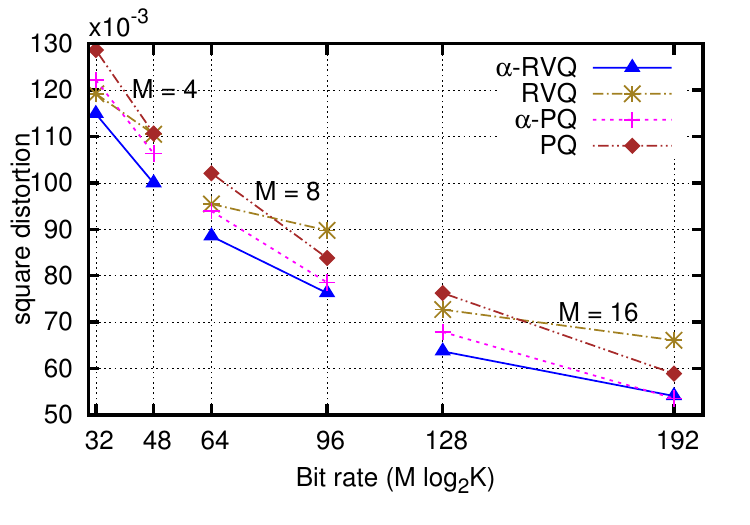}&
		\includegraphics[trim = 0mm 3mm 0mm 1mm, clip, width=.33\textwidth]{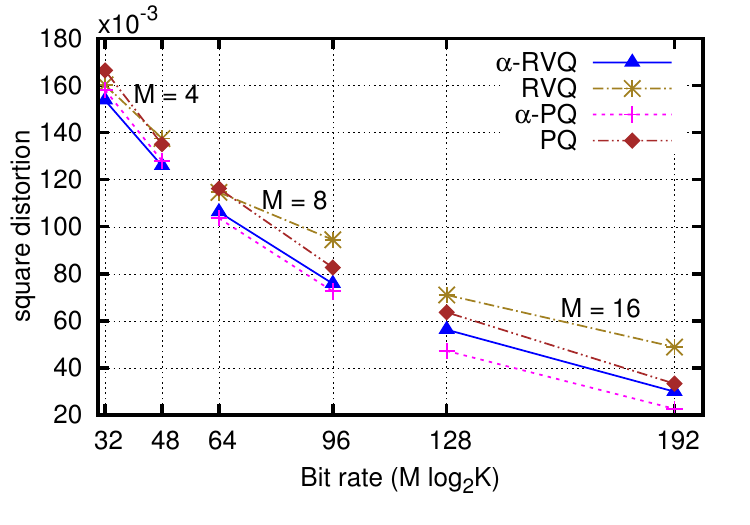}\\
		SIFT1M & GIST1M & VLAD500K	
	\end{tabular}
	\caption{{\bf Accuracy of structured encoding, with and without coefficients}. Squared reconstruction errors produced by structured encoding (PQ and RVQ) and proposed sparse encoding extensions (\aPQ~ and \aRQ). For each method, $M=4,8,16$ and $\log_2 K =8,12$ are reported. In absence of coefficient quantization here, each code has $M\log_2 K$ bits, \ie\, 64 bits for $(M,K)=(8,256)$.}
	\label{fig:dst}
\end{figure*}

\noindent\textbf{Introducing coefficients}\quad Before moving to the main experiments, we first investigate how the key idea of including scalar coefficients into structured quantization allows more accurate vector encoding. To this end, we compare average reconstruction errors, $\frac{1}{N}\sum_{\bx\in\cX}\|\bx-Q(\bx)\|^2_2$, obtained on the different datasets by RVQ (resp. PQ) and the proposed approach \textit{before vector quantization of coefficient vector}, which we denote \aRQ~(resp. \aPQ), see Fig. \ref{fig:dst}. Three structure granularities are considered, $M=4,~8$ and $16$. Note that in RVQ and \aRQ, increasing the number of layers from $M$ to $M'>M$ simply amounts to resuming recursive encoding of residuals. For PQ and \aPQ~ however, it means considering two different partitions of the input vectors: the underlying codebooks/dictionaries and  the resulting encodings have nothing in common. 

Reconstruction errors (distortions) are also reported for $K=2^8$ and $2^{12}$ respectively.
For a given method, reconstruction error decreases when $M$ or $K$ increases. Also, as expected, \aRQ~ (resp. \aPQ) is more accurate than RVQ (resp. PQ) for the same $(M,K)$. As we shall see next, most of this accuracy gain is retained after quantizing, even quite coarsely, the coefficient vectors. \medskip

\begin{figure*}
	\scriptsize
\begin{tabular}{ccc}
\includegraphics[trim = 0mm 1mm 0mm 1mm, clip, width=.33\textwidth]{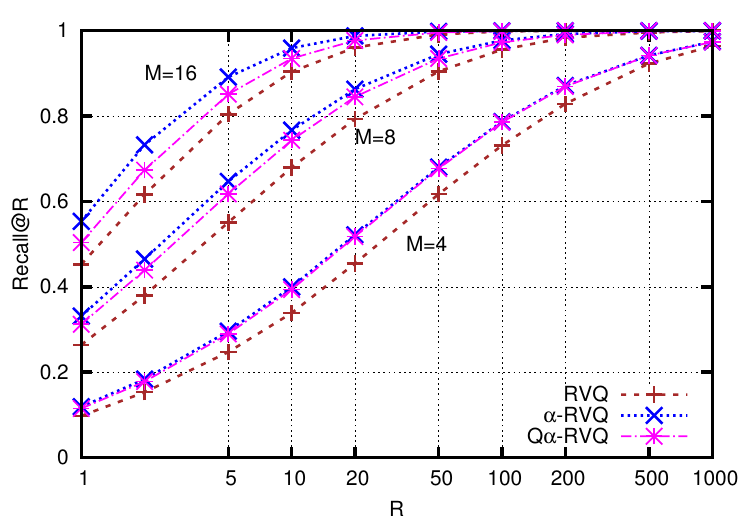} &
\includegraphics[trim = 0mm 1mm 0mm 1mm, clip, width=.33\textwidth]{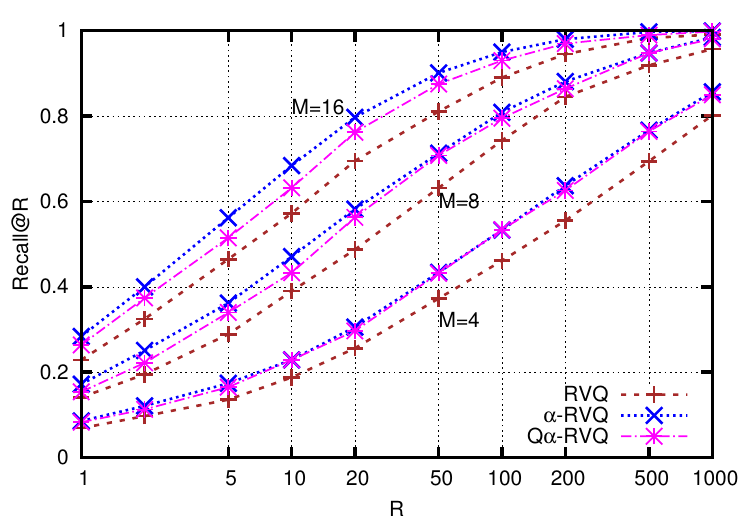} &
\includegraphics[trim = 0mm 1mm 0mm 1mm, clip, width=.33\textwidth]{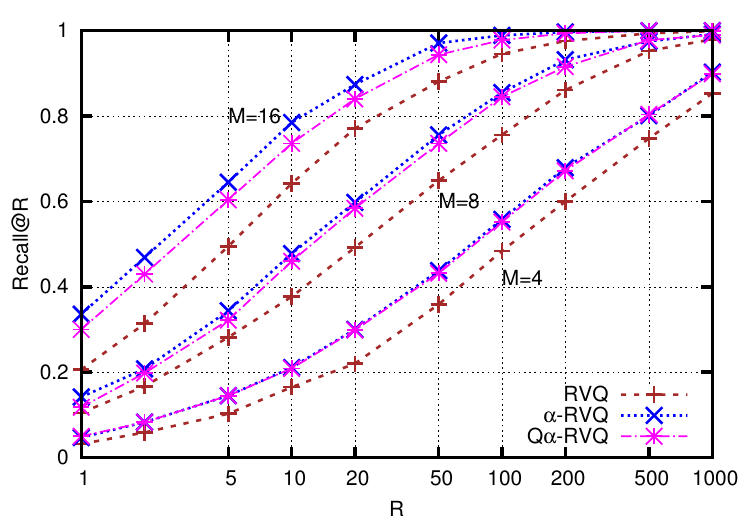}\\
\includegraphics[trim = 0mm 1mm 0mm 1mm, clip, width=.33\textwidth]{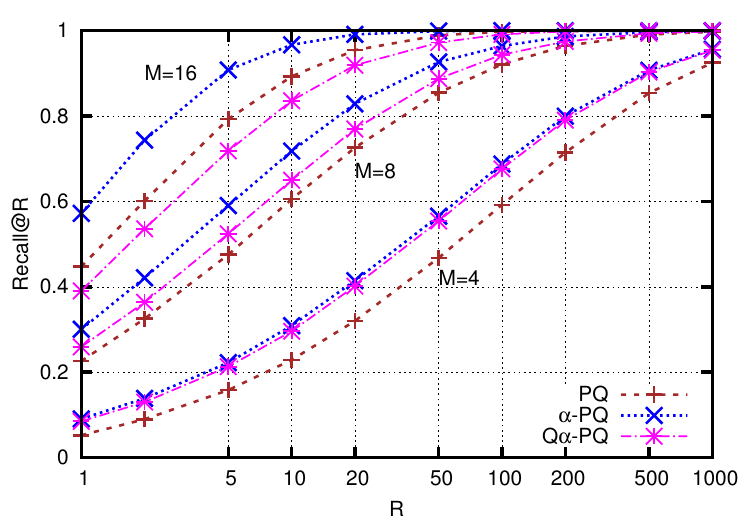} &
\includegraphics[trim = 0mm 1mm 0mm 1mm, clip, width=.33\textwidth]{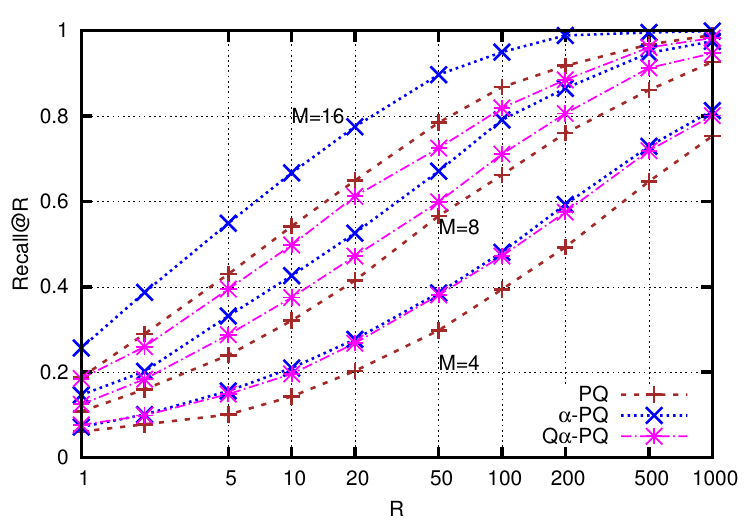} &
\includegraphics[trim = 0mm 1mm 0mm 1mm, clip, width=.33\textwidth]{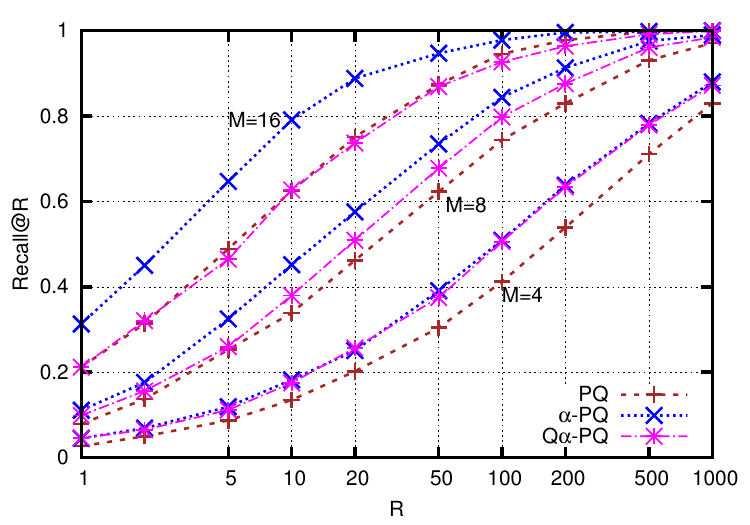}\\
SIFT1M & GIST1M & VLAD500K
\end{tabular}
\caption{{\bf Impact of 1-byte $\balpha$ quantization on performance}. Recall@{\tt R} curves for \qaRQ, \aRQ~and RVQ (resp. \qaPQ, \aPQ~and PQ) on the three datasets, with $M\in\{4,8,16\}$, $K=256$ and $P=256$.}
	\label{fig:r@r_qapq}
\end{figure*}

\noindent\textbf{Quantizing coefficients}\quad
Figure~\ref{fig:r@r_qapq} shows the effect of this quantization on the performance, in comparison to no quantization (sparse encoding) and to classic structured quantization without coefficients. For these plots, we have used one byte encoding for $\balpha$, \ie, $P=256$, along with $M\in\{4, 8, 16\}$ and $K=256$.
With this setting, \qaRQ~ (resp. \qaPQ) is compared to both \aRQ~and RVQ (resp. \aPQ~and PQ) with the same values of $M$ and $K$.  This means in particular that \qaRQ~ (resp. \qaPQ) benefits from one extra byte compared to RVQ (resp. PQ).  
Note that allowing one more byte for RVQ/PQ encoding would significantly increase its learning, encoding and search practical complexities.

Since $\balpha$ has $M$ dimensions, its quantization with a single byte gets cruder as $M$ increases, leading to a larger relative loss of performance as compared to no quantization. For $M=4$, one byte quantization suffices in both structures to almost match the good performance of unquantized sparse representation. For $M=16$, the increased degradation remains small within \qaRQ. However it is important with \qaPQ: for a small budget allocated to the quantization of $\balpha$, it is even outperformed by the PQ baseline. This observation is counter-intuitive (with additional information, there is a loss). The reason is that the assignment is greedy: while the weights are better approximated w.r.t. a square loss, the vector reconstruction is inferior with Eqn~\ref{eq:coding}. A non-greedy exploration strategy as in AQ would address this problem but would also dramatically increase the assignment cost. 
This suggests that the size $P$ of the codebook associated with $\balpha$ should be adapted to the number $M$ of layers.

Hereafter, we measure for each dataset the minimum number of bits that must be dedicated to coefficients quantization ($\log_2 P$) to ensure that the reconstruction error with structured sparse coding remains below the one of the corresponding structured quantization method. The results are as follows (SIFT/GIST/VLAD):
\begin{center}{\small
 \begin{tabular}{lcc}
 		& \qaRQ~ vs. RVQ & \qaPQ~ vs. PQ 	\\ \hline\hline
 $M=4$ 	& $1/1/1$ 		& $3/3/2$			\\ \hline
 $M=8$ 	& $1/1/1$		& $6/5/5$			\\ \hline
 $M=16$ & $1/1/1$ 		& $15/11/12$		\\ \hline
 \end{tabular}}
\end{center}
Interestingly, the first bit allocated to $\balpha$ improve upon RVQ for all the settings and datasets. In contrast and as discussed before, for \qaPQ, more bits must be allocated to the weights for larger $M$ to guarantee a better representation. For instance, an overhead of 6 bits is required for $M=8$. \medskip

\noindent\textbf{Comparing at fixed code size}\quad
For large scale search, considering (almost) equal encoding sizes is a good footing for comparisons. This can be achieved in different ways. In the case of RVQ and \qaRQ, the recursive nature of encoding provides a natural way to allocate the same encoding budget for the two approaches: we compare \qaRQ~with $(M,K,P)$ to RVQ with $M$ codebooks of size $K$ and a last one of size $P$.
For PQ and \qaPQ, things are less simple: adding one codebook to PQ to match the code size of \qaPQ~ leads to a completely different partition of vectors, creating new possible sources of behavior discrepancies between the two compared methods. Instead, we compare PQ with $M$ codebooks of size $K$ to \qaPQ~ with $M$ dictionaries of size $K/2$ and $P=2^M$ codewords for coefficient vectors. This way, vector partitions are the same for both, as well as the corresponding code sizes ($M \log_2 K$ bits for PQ and $M\log_2 \frac{K}{2} + \log_2 2^M = M \log_2 K$ bits for \qaPQ).

Sticking to these rules, we shall compare next structured quantization and quantized sparse representation for equal encoding sizes. \medskip

\begin{figure*}
	\centering
	\scriptsize
	\begin{tabular}{ccc}	
\includegraphics[trim = 0mm 1mm 0mm 3mm, clip, width=.33\textwidth]{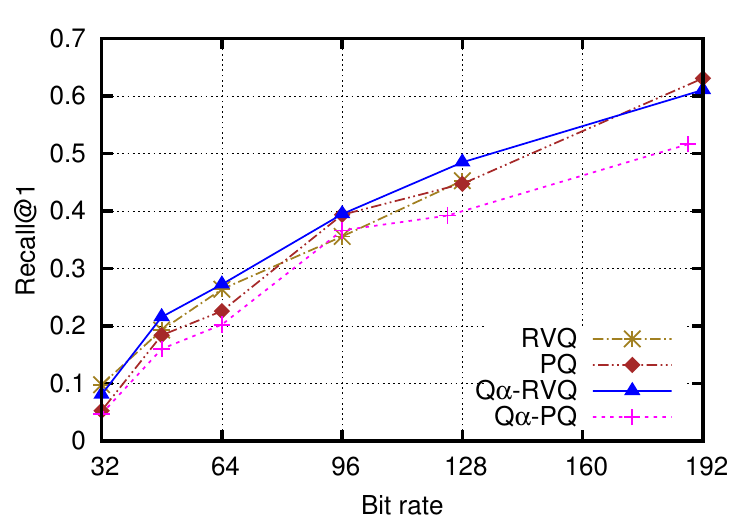}&
\includegraphics[trim = 0mm 1mm 0mm 3mm, clip, width=.33\textwidth]{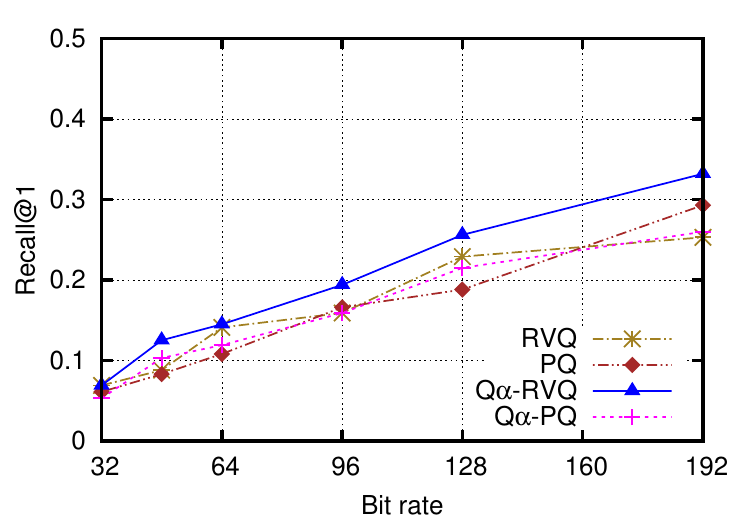}&
\includegraphics[trim = 0mm 1mm 0mm 3mm, clip, width=.33\textwidth]{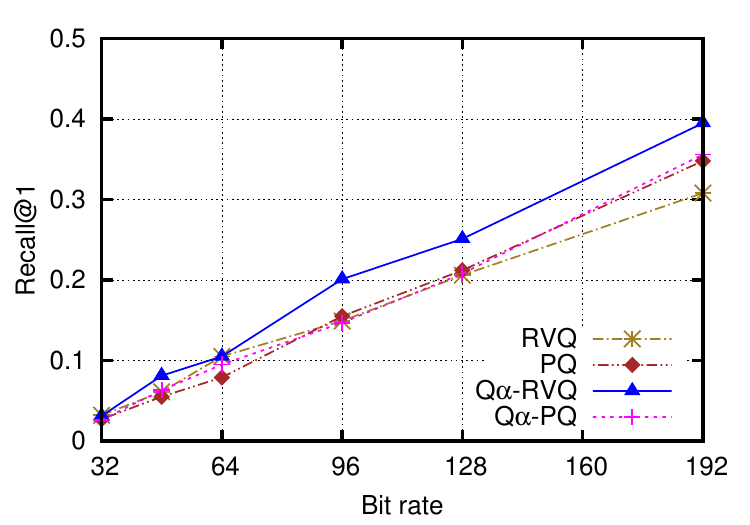}\\
SIFT1M & GIST1M & VLAD500K	
	\end{tabular}
	\caption{{\bf Comparative CS-aNN performance for different encoding sizes}. Recall@1 on the three datasets for increasing number of encoding bits, comparing PQ and RVQ with \qaPQ~ and \qaRQ~ respectively.
	}
	\label{fig:r@1}
\end{figure*}

\noindent\textbf{CS-aNN}\quad We compare RVQ to \qaRQ~ and PQ to \qaPQ~ for different code sizes, from 8 to 24 bytes per vector, on the task of maximum cosine similarity over $\ell^2$-normalized vectors. Corresponding Recall@1 curves are in Fig. \ref{fig:r@1}.
\qaRQ~ clearly outperforms RVQ on all datasets, even with a substantial margin on GIST1M and VLAD500K, \ie, around 30\% relative gain at 24 bytes.
The comparison between PQ and \qaPQ~ leads to mixed conclusions: while \qaPQ~is below PQ on SIFT1M, it is slightly above for GIST1M and almost similar for VLAD500K. Note however that, for the same number $M \log_2 K$ of encoding bits, \qaPQ~ uses $M\frac{K}{2}+2^M$ centroids, which is nearly half the number $MK$ of centroids used by PQ in low $M$ regimes (\eg, when $K=256$, 528 vs. 1024 centroids for $M=4$ and 1280 vs. 2048 centroids for $M=8$). Much fewer centroids for equal code size and similar performance yield computational savings in learning and encoding phases. \medskip

\noindent\textbf{Euclidean aNN}\quad In order to conduct comparison with other state-of-art methods such as extensions of PQ and of RVQ, we also considered the Euclidean aNN search problem, with no prior normalization of vectors. For this problem, the proposed approach applies similarly since the minimization problem $\argmin_{\bx\in\cX} \|\by - Q(\bx)\|^2 = \argmax_{\bx\in\cX} \trsp{\by}Q(\bx) - \frac{\|Q(\bx)\|^2}{2}$
involves the same quantities as the one in (\ref{eq:cs_ann}).

\begin{figure*}
	\centerline{\scriptsize
	\begin{tabular}{ccc}	
	\includegraphics[trim = 0mm 1mm 0mm 3mm, clip, width=.25\textwidth]{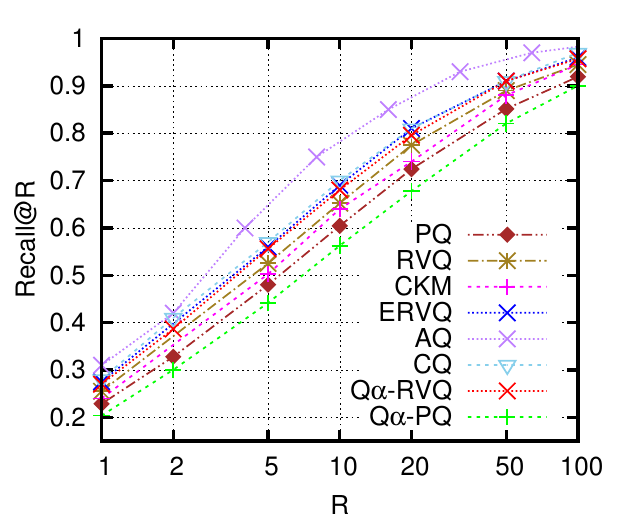}
	\includegraphics[trim = 0mm 1mm 0mm 3mm, clip, width=.25\textwidth]{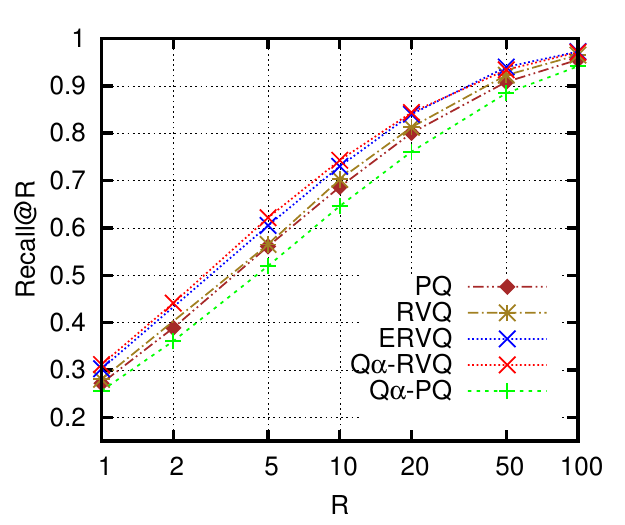}& \quad &
	\includegraphics[trim = 0mm 1mm 0mm 3mm, clip, width=.25\textwidth]{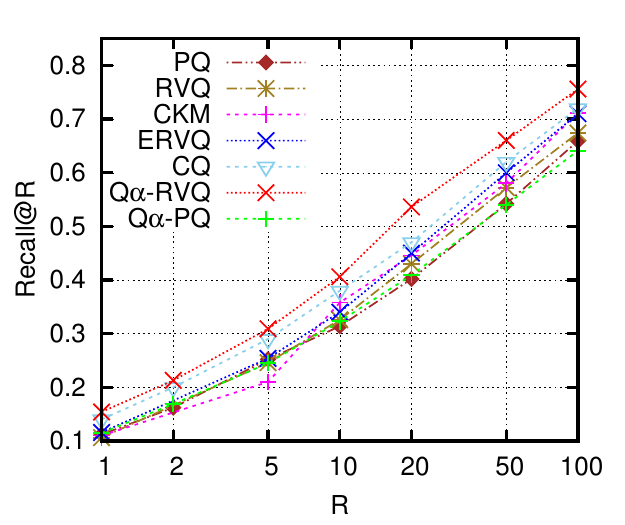}
	\includegraphics[trim = 0mm 1mm 0mm 3mm, clip, width=.25\textwidth]{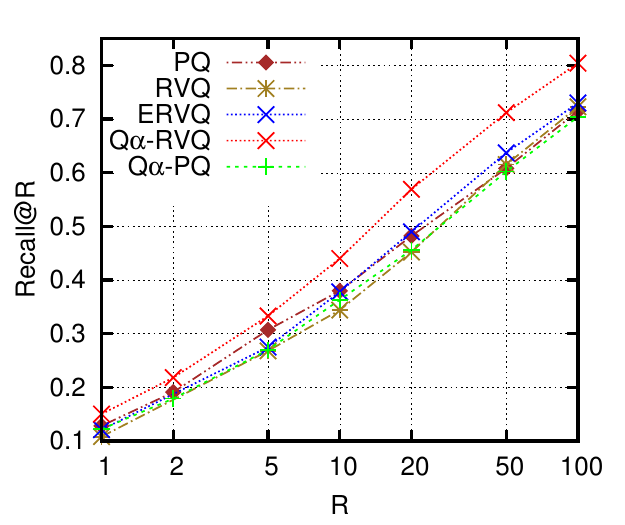}\\
	SIFT1M: 64 bits and 72 bits encoding& & GIST1M: 64 bits and 72 bits encoding
	\end{tabular}}
	\caption{{\bf Performance comparison for Euclidean-aNN}. Recall@{\tt R} curves on SIFT1M and GIST1M, comparing proposed methods to PQ, RVQ and to some of their extensions, CKM~\cite{NF13}, ERVQ~\cite{AYWHG15}, AQ~\cite{BL14} and CQ~\cite{ZDW14}.
	} 
	\label{fig:l2_recall}
\end{figure*}

Recall@{\tt R} curves are provided in Fig. \ref{fig:l2_recall} on two of the three datasets, relying on results reported in \cite{NF13} for CKM, RVQ and ERVQ, and \cite{BL14}, \cite{ZDW14} for AQ and CQ respectively. We observe again that $\qaPQ$ is below PQ on SIFT but on par with it on GIST.
On SIFT, $\qaRQ$, ERVQ and CQ perform similarly, while on GIST $\qaRQ$ outperforms all, including CQ and ERVQ. As discussed in Section \ref{sec:related}, AQ is the most accurate but has very high encoding complexity. CQ also has higher encoding complexity compared to our simple and greedy approach. 

Table \ref{tab:dist_unnorm} shows reconstruction errors for the same setting as in Fig. \ref{fig:l2_recall}. This is consistent with the results in Fig. \ref{fig:l2_recall}, and shows again that $\qaRQ$ is the best performer and that $\qaPQ$ does not improve on PQ with the same number of encoding bits.

\begin{table}
	\centering
\begin{tabular}{ccccc}
	\hline
	& \multicolumn{2}{c}{SIFT}	&	\multicolumn{2}{c}{GIST}					\\ \hline
	encoding bits 	& 64 			& 72 			& 64 			& 72  	\\ \hline\hline
	PQ 				& 23515			& 20054			& 0.7121 		& 0.6733 	\\ \hline
	\qaPQ 			& 25859			& 22007			& 0.7224 		& 0.6868	\\ \hline
	RVQ 			& 22170			& 20606	 		& 0.6986 		& 0.6734 	\\ \hline
	\qaRQ 			& \textbf{22053}& \textbf{19976}& \textbf{0.6537}& \textbf{0.6174} \\ \hline
\end{tabular}			 
\caption{{\bf Comparative distortions in Euclidean aNN setting}. Average squared reconstruction errors on un-normalized SIFT1M and GIST1M.}
\label{tab:dist_unnorm}
\end{table}

Note that the lower performance of $\qaPQ$ compared to PQ is because it uses half the number of codewords to have equal or smaller memory footprint. Relative timings in Tab.~\ref{tab:timings} indicate $\qaPQ$ is substantially faster for learning and encoding in this setting. Our methods are slower in search but this overhead has minimal effect in the applications with very large scale data as we shall see in our billion-scale experiments. 

\begin{table}
\centering
	\begin{tabular}{llllr}
	\hline
	& PQ & $\qaPQ$ & RVQ & $\qaRQ$ \\ \hline \hline
	learn	& 1	&	0.212	&	1.250	&	0.719	\\ \hline
	encode	& 1	&	0.206	&	1.347	&	0.613	\\ \hline 
	search	& 1	&	1.867	&	1.220	&	1.909	\\ \hline
	\end{tabular}\hfil
	\caption{{\bf Relative timings}. Learning, encoding and search timings w.r.t PQ on SIFT1M with 64 bits encoding. \qaPQ~and \qaRQ~ have faster learning/encoding as they use inner product instead of $\ell^2$ distance and have fewer codewords. 
	}\label{tab:timings}
\end{table}

Table~\ref{tab:pq_methods} provides recall rates for various PQ based methods on SIFT1M with 64 bits encoding. CKM and OPQ are very similar extensions on PQ and thus perform similarly. The improvement provided by CKM/OPQ on PQ is complimentary to $\qaPQ$. By using OPQ instead of PQ within $\qaPQ$, calling it \mbox{Q$\alpha$-OPQ}, we get similar gains as OPQ gives over PQ. This can be seen by comparing the gains of \mbox{Q$\alpha$-OPQ} over \qaPQ~and OPQ over PQ. These results of OPQ and \mbox{Q$\alpha$-OPQ} are not plotted in \ref{fig:l2_recall} to maintain clarity.

\begin{table}
	\centering
\begin{tabular}{crrrrr}
\hline
	Recall~  & PQ~ & CKM~ & OPQ~ & \qaPQ~ & \mbox{Q$\alpha$-OPQ}	\\ \hline \hline
	$R@1$   & 0.228 & 0.240 & 0.243	& 0.204 &  0.227  \\ \hline
	$R@10$  & 0.604 & 0.640	& 0.638	& 0.562 &  0.603  \\ \hline
	$R@100$ & 0.919 & 0.945	& 0.942	& 0.900 &  0.927  \\ \hline
\end{tabular}
	\caption{{{\bf CKM/OPQ comparison with PQ}. Performance of PQ based methods on SIFT1M with 64 bits encoding. $(M, K) = (8, 256)$ for PQ, CKM and OPQ and $(M, K, P) = (8, 128, 256)$ for our methods.}
	 \label{tab:pq_methods}}
\end{table}

\begin{figure*}
\begin{tabular}{c}
\raisebox{-0.5\height}{\includegraphics[width=.5\textwidth]{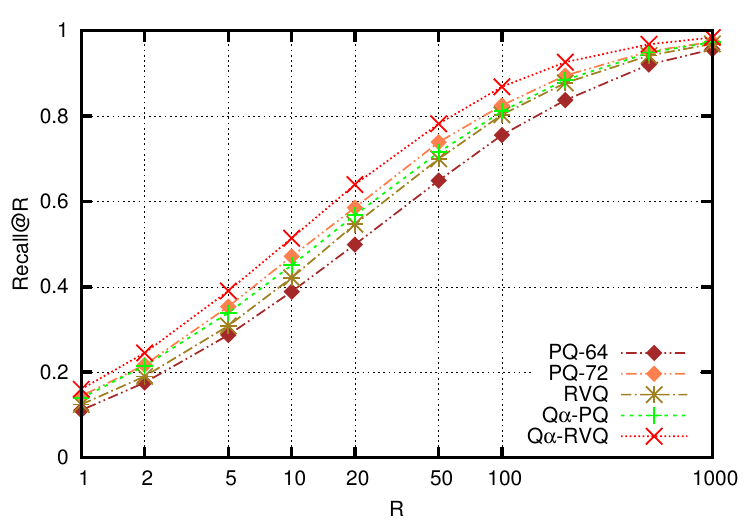}}
\quad\begin{tabular}{llllr}
\toprule
Method (\textit{b})	& R@1 	& R@10	& R@100	& time  \\ \toprule
PQ-64 (8)			& 0.111	& 0.388	& 0.756	& 1.00 	\\ \hline 
PQ-72 (9)			& 0.144	& 0.471	& 0.825	& 1.03	\\ \hline 
RVQ (9)				& 0.124	& 0.421	& 0.803	& 1.02 	\\ \hline 
\qaPQ (9)			& 0.139	& 0.450	& 0.811	& 1.69 	\\ \hline 
\qaRQ (10)			& 0.160	& 0.514	& 0.868	& 1.72 	\\ \hline 
\qaRQ$_{128}$		& 0.160 & 0.514 & 0.868 & 0.89  \\ \hline 
\qaRQ$_{8}$ & 0.151 & 0.467 & 0.730 & 0.17 \\ \hline 
\end{tabular}
\end{tabular}	
\caption{{\bf Large scale performance with IVF}. Recall@{\tt R} on the BIGANN 1B-SIFT dataset and 10K queries. For all methods, $M=8$ and $K=256$, except for ``PQ-72" ($K=512$). For quantized sparse coding methods, $P=256$ and norms in residual variant are quantized over 256 scalar values, resulting encoding sizes (\textit{b}) being given in bytes per vector. All methods share the same IVF index with $K_c=2^{13}$ and $W_c=64$. Subscripted \qaRQ~denotes variants with additional pruning ($W'_c=128$ and $8$ resp.). Search timings are expressed relative to PQ-64.} 
\label{fig:bigann_recall}
\end{figure*}

\noindent\textbf{Very large scale experiments on BIGANN}\quad We validate our approach on large scale experiments carried out on the BIGANN dataset \cite{jegou2011searching}, which contains 1 billion SIFT vectors ($N=1B$, $R=1M$ out of the original 100M training set and $S=10K$ queries). At that scale, an inverted file (IVF) system based on a preliminary coarse quantization of vectors is required. In our experiments, each vector is quantized over $K_c=8192$ centroids, and it is its residual relative to assigned centroid that is fed to the chosen encoder. At search time, the query is multiply assigned to its $W_c=64$ closest centroids and $W_c$ searches are conducted over the corresponding vector lists (each of average size $N/K_c$).
Performance is reported in Fig. \ref{fig:bigann_recall} for PQ, RVQ and their proposed extensions. For all of them the setting is $M=8$ and $K=256$, except for PQ-72 bits ($K=512$). All of them use the exact same IVF structure, which occupies approximately 4GB in memory (4B per vector). For RVQ and \qaRQ, norms of approximated database vectors are quantized over 256 scalar values. 

The best performance is obtained with the proposed \qaRQ~approach, which requires 10 bytes per vector,
thus a total of 14GB for the whole index. The second best aNN search method is PQ-72 bits, which requires 9 bytes per vector, hence 13GB of index. While both indexes have similar sizes and fit easily in main memory, PQ-72 relies on twice as many vector centroids which makes learning and encoding more expensive. 

The superior performance of \qaRQ~ comes at the price of a 70\% increase of search time per query compared to PQ. This can nonetheless be completely 
compensated for since the hierarchical structure of \qaRQ~ lends itself to additional pruning after the one with IVF. The $W'_c$ atoms most correlated with the query residual in $C^1$ are determined, and dataset vectors whose first layer encoding uses none of them are ignored. For $W'_c =128$, search time is reduced substantially, making \qaRQ~10\% faster than PQ, with no performance loss (hence superior to PQ-72). A more drastic pruning ($W'_c=8$) makes performance drop below that of PQ-72, leaving it on par with PQ-64 while being almost 6 times faster.

A variant of IVF, called ``inverted multi-index'' (IMI) \cite{BL12} is reported to outperform IVF in speed and accuracy, by using two-fold product quantization instead of vector quantization to produce the first coarse encoding. Using two codebooks of size $K_c$, one for each half of the vectors, IMI produces  $K_c^2$ inverted lists. We have run experiments with this alternative inverted file system, using $K_c=2^{14}$ and scanning a list of $T = 100K,~30K$ or $10K$ vectors, as proposed in \cite{BL12}. The comparisons with PQ-64 based on the same IMI are summarized in Tab. \ref{tab:imi_bigann} in terms of recall rates and timings. For all values of $T$, the proposed $\qaRQ$ and $\qaPQ$ perform the best and with similar search time as RVQ and PQ-64. 
Also, $\qaRQ$ with $T=30K$ has the same recall@100 as PQ-64 with $T=100K$ while being twice as fast (14ms vs. 29ms per query). For a fixed $T$, PQ-64  and $\qaPQ$ (resp. RVQ and $\qaRQ$) have the same search speed, as the overhead of finding the $T$ candidates and computing look-up tables dominates for such relatively short lists. The $T$ candidates for distance computation are very finely and scarcely chosen. Therefore, increasing the size $K$ of dictionaries/codebooks in the encoding method directly affects search time. This advocates for our methods, as for equal $(M,K)$ and an extra byte for encoding coefficients, $\qaRQ$ and $\qaPQ$ always give better performance. Compared to PQ-72, $\qaPQ$~ is faster (only half the number of codewords is required in the quantization) with slightly lower accuracy. $\qaRQ$ is more accurate with extra execution time compared to PQ-72.

\begin{table*}
\centerline{\footnotesize
\begin{tabular}{l cccc c cccc c cccc}
\toprule
 & \multicolumn{4}{c}{$T=$100K} & \quad & \multicolumn{4}{c}{$T=$30K} & \quad & \multicolumn{4}{c}{$T=$10K} \\
        	\cline{2-5}\cline{7-10} \cline{12-15}
        	Method (\textit{b}) & R@1 & R@10 & R@100 & time & \quad & R@1 & R@10 & R@100 & time & \quad & R@1 & R@10 & R@100 & time \\
        	\hline
        	PQ-64 (8) &	0.170 & 0.535 &	0.869 & 29 & \quad & 0.170 & 0.526 & 0.823 & 11  & \quad & 0.166 & 0.495 & 0.725 & 5  \\
        	RVQ (9) & 0.181 & 0.553 & 0.877 & 37  & \quad & 0.180	& 0.542	& 0.831 & 14 & \quad  & 0.174 & 0.506 & 0.729 & 8  \\
        	\qaPQ (9) & 0.200 & 0.587 & 0.898 & 30 & \quad  & 0.198 & 0.572 & 0.848 & 11  & \quad & 0.193 & 0.533 & 0.740 & 5  \\
			\qaRQ(10) & 0.227 & 0.630	& 0.920 & 37 & \quad  & 0.225 & 0.613 & 0.862 & 14  & \quad &	0.217 & 0.566 & 0.747 & 8  \\
			PQ-72 (9) &	0.207 & 0.603 & 0.902 & 34 & \quad & 0.205 & 0.586 & 0.849 & 12  & \quad & 0.2 & 0.547 & 0.739 & 6  \\
        	\bottomrule
\end{tabular}}
\caption{{\bf Performance and timings with IMI on 1B SIFTs}. Recalls are reported along with search time in milliseconds per query as a function of the length $T$ of candidate list to be exhaustively scanned. For each method, the encoding size (\textit{b}) is given in bytes per vector.}
\label{tab:imi_bigann}
\end{table*}

\section{Discussion and conclusion}\label{sec:conclusion}

In this work we present a novel quantized sparse representation that is specially designed for large scale approximate nearest neighbour search. The residual form of this representation, \qaRQ, clearly outperforms RVQ in all datasets and settings, for equal code size. Within the recursive structure of residual quantization, the introduction of additional coefficients in the representation thus offers accuracy improvements that translate into aNN performance gains, even after drastic vector quantization of these coefficients. Interestingly, the gain is much larger for image level descriptors (GIST and VLAD) which are key to very large visual search. One possible reason for the proposed approach to be especially successful in its residual form lies in the rapid decay of the coefficients that the hierarchical structure induces. This facilitates quantization of coefficient vectors, even with 1 byte only. 
In its partitioned variant, this property is not true anymore, and the other proposed approach, \qaPQ, brings less gain. It does however improve over PQ for image-level descriptors, especially in small $M$ regimes, while using fewer centroids.

As demonstrated on the billion-size BIGANN dataset, the proposed framework can be combined with existing inverted file systems like IVF or IMI to provide highly competitive performance on large scale search problems. In this context, we show in particular that both \qaPQ~and \qaRQ~offer higher levels of search quality compared to PQ and RVQ for similar speed and that they allow faster search with similar quality.
Regarding \qaRQ, it is also worth noting that its hierarchical structure allows one to prune out most distant vectors based only on truncated descriptors, as demonstrated on BIGANN1B within IVF system. Conversely, this nested structure permits to refine encoding if desired, with no need to retrain and recompute the encoding up to the current layer. 

On a different note, the successful deployment of the proposed quantized sparse encoding over million to billion-sized vector collections suggests it could help scaling up sparse coding massively in other applications.